\documentclass{article}
\usepackage{ijcai26}
\usepackage{enumitem}

\usepackage{times}
\usepackage{soul}
\usepackage{url}
\usepackage[hidelinks]{hyperref}
\usepackage[utf8]{inputenc}
\usepackage[small]{caption}
\usepackage{graphicx}
\usepackage{amsmath}
\usepackage{amsthm}
\usepackage{amsfonts}

\usepackage{booktabs}
\usepackage{algorithm}
\usepackage{algorithmic}
\usepackage[switch]{lineno}
\usepackage{placeins}
\usepackage{tabularx}  % For flexible width tables

\title{AlgoEvolve: LLM-driven Meta-evolution of Algorithmic Trading Programs}

\author{
Dhruv Sharma$^1$
\and
Dr. Gautam Shroff$^2$\and
\affiliations
$^1$Indraprastha Institute of Information Technology, Delhi\\
$^2$Indraprastha Institute of Information Technology, Delhi\\
\emails
\{dhruv22170, gautam.shroff\}@iiitd.ac.in,
}

\begin{document}

\maketitle

\begin{abstract}
Recent work shows that Large Language Models (LLMs) can act as semantic mutation operators for the evolutionary discovery of programs and proofs. Most current applications focus on static coding benchmarks. We extend this paradigm to algorithmic trading. This domain is uniquely challenging because it is noisy, non-stationary, and highly discontinuous. We present AlgoEvolve, an LLM-driven evolutionary framework that generates, evaluates, and iteratively improves executable trading strategies. These strategies are expressed as Python code and evaluated through a rigorous testing protocol. Across multiple experiments, the system exhibits emergent regime-adaptive strategy logic, including autonomous shifts in trading rules. We further introduce a meta-evolutionary outer loop that evolves the prompts guiding program synthesis in the inner loop. This outer loop discovers improved search heuristics. These heuristics balance exploration and exploitation while reducing zero-trade failures. They consistently outperform initial human-designed instructions. The results demonstrate that LLM-based semantic evolution provides a viable approach for continual program synthesis in complex environments.
\end{abstract}

\section{Introduction}

Procedures that aim to discover optimal strategies for algorithmic trading in financial markets, e.g., using machine-learning, need to synthesize and select heterogeneous market signals as well as adapt to non-stationary regimes \cite{hambly2023recent,yu2025fincon,li2023trading}. A trading strategy's performance (i.e., profit and loss) is most often a non-differentiable and highly discontinuous function of any parameters it may choose to use (e.g., thresholds, model weights etc.) \cite{zhang2020deep}, due to the noisy character of the domain, i.e., a very low signal-to-noise ratio \cite{lim2021time}. Traditional deep learning and reinforcement learning (RL) approaches are also limited by their reliance on 'black-box' parametric optimization, lacking the transparency often required by regulatory frameworks \cite{arsenault2025xai}.
These are also prone to overfitting to historical noise \cite{de2018advances}, leading to severe degradation during abrupt regime shifts.

Recently, Large Language Models (LLMs) have demonstrated their potential in financial decision-making and addressing the  limitations of parametric models: Beyond serving as multi-modal feature extractors \cite{chen2021decision,yang2023fingpt}, LLMs exhibit emergent capabilities in capturing long-range temporal dependencies and generalizing across diverse market regimes through in-context learning \cite{jin2023time,yu2024finmem}. However, while these models are being increasingly used as "agentic" controllers, they have primarily been evaluated as one-shot generators or static predictors. We propose using LLMs as semantic mutation operators capable of iterative program refinement. By drawing inspiration from recent breakthroughs in symbolic discovery such as FunSearch \cite{romera2023mathematical} and AlphaEvolve \cite{novikov2025alphaevolve}, as well as the scaling of test-time compute through iterative editing \cite{ehrlich2025codemonkeys}, we demonstrate that LLMs can participate in an iterative evolutionary discovery process, generating and continually improving executable algorithmic trading strategies.

\begin{figure*}[t]
    \centering
    \includegraphics[width=0.95\textwidth]{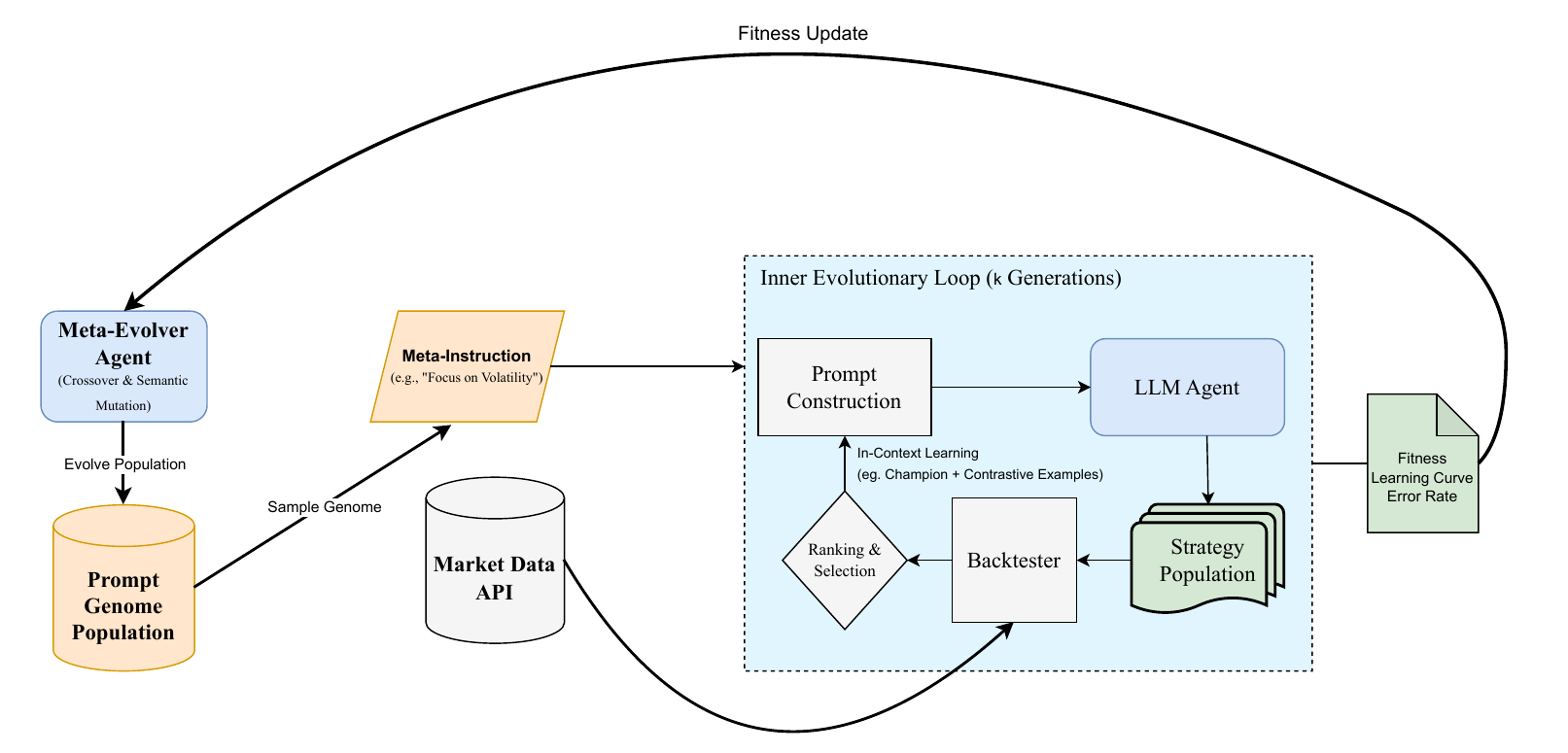}
    \caption{\textbf{The AlgoEvolve Framework.} Our hierarchical architecture co-evolves symbolic trading policies and their discovery heuristics. (A) The Inner Loop utilizes an LLM as a semantic mutation operator to iteratively refine executable Python strategies, evaluated via a rigorous walk-forward protocol. (B) The Outer Loop performs meta-evolution on the Prompt Genome, discovering superior search instructions that adapt to market non-stationarity and stabilize the discovery process against common failure modes like "zero-trade" stagnation.}
    \label{fig:architecture}
\end{figure*}

We introduce \textbf{AlgoEvolve}, a hierarchical framework where the LLM functions as a semantic mutation operator to discover executable trading policies. Building upon the "coding agent" paradigm established by AlphaEvolve \cite{novikov2025alphaevolve}, which orchestrates LLM pipelines for scientific discovery, we propose a specialized bi-level meta-evolutionary architecture: An inner loop evolves strategies using LLMs provided with a search prompt, strategy code, and recent performance data. Additionally, an outer loop evolves the search prompt itself. Unlike existing approaches that rely on static instructions or unstructured feedback \cite{fernando2023,yang2023opro}, AlgoEvolve treats the prompt as a structured \textit{prompt genome}, enabling the system to autonomously discover and refine its own discovery heuristics across shifting market regimes.

As AlgoEvolve is run continually and iteratively over time, it exhibits qualitative shifts in strategy logic that suggest a departure from simple historical fitting. We observe autonomous transitions in trading paradigms, where the system independently moves beyond human-provided trend-following priors, discovering complex, regime-adaptive rules such as multi-factor scoring and price-action heuristics. These discovered strategies frequently utilize multi-bar patterns that mirror human-engineered structures, yet are optimized for the specific volatility profiles of the target assets. At the outer level, the meta-evolutionary layer discovers \textit{evolved prompts} that stabilize reasoning and effectively mitigate the 'zero-trade' failure modes observed in static LLM-based search.

Our contributions are as follows: \textbf{(1)} we introduce an end-to-end LLM-driven evolutionary framework for discovering and improving algorithmic trading strategies, resulting in an annualized Sharpe ratio of 5.60; \textbf{(2)} we demonstrate that AlgoEvolve independently abandons human-designed trend-following priors to discover superior, regime-adaptive logic; \textbf{(3)} we propose a meta-evolutionary Outer Loop that evolves the inner loop's Evolver Prompt itself, enabling autonomous refinement of the search heuristic; and \textbf{(4)} we submit that LLM-based semantic evolution is a viable approach for program synthesis in noisy, non-differentiable, and high-dimensional environments such as algorithmic trading.

\section{Related Work}

\textbf{LLM-Driven Program Synthesis}
Program synthesis has increasingly expanded from one-shot autoregressive generation \cite{nijkamp2023codegen}
to iterative discovery, where LLMs act as semantically informed mutation operators within evolutionary architectures
\cite{romera2023mathematical}. To manage vast program search spaces, current research integrates LLM reasoning
with formal constraints, including syntactic guidance for enumerative synthesis \cite{li2024} and grammar-based
structural constraints in hybrid human–LLM workflows \cite{barke2022grounded}. Performance in these frameworks
is increasingly driven by test-time compute, leveraging iterative editing \cite{ehrlich2025codemonkeys},
population-based exploration over program proposals \cite{real2020automlzero}, and execution feedback cycles
\cite{yang2023opro} to resolve complex logic. Unlike classical Genetic Programming, LLM-driven evolution better
captures semantic intent required for high-dimensional optimization \cite{fernando2023}, providing a foundation
for synthesis in noisy, discontinuous environments such as algorithmic trading.

\textbf{LLMs for Financial Decision Making}
Large language models (LLMs) have been increasingly applied to financial tasks, ranging from sentiment analysis and domain-specific modeling \cite{araci2019,wu2023bloomberggpt} to agentic trading systems that perform inference-time decision making \cite{yu2025fincon,zhang2024finagent,wu2025mountainlion,song2025tradeinminutes}. Recent frameworks further integrate reinforcement learning \cite{xiong2025flag} or internal competition mechanisms \cite{zhao2025contesttrade} to improve robustness in noisy markets, but rely on continuous model inference during deployment.

In contrast, \emph{AlgoEvolve} employs LLMs only at design time to synthesize explicit, executable trading strategies. By evolving symbolic Python programs rather than querying an LLM at runtime, the framework achieves zero inference-time latency, intrinsic interpretability, and improved robustness to non-stationary market regimes. This formulation reframes LLMs as symbolic strategy designers rather than real-time trading agents, better aligning language-based reasoning with the operational constraints of algorithmic trading.

\textbf{Evolutionary Computation for Financial Strategy Discovery}
Evolutionary methods for trading range from Genetic Programming for symbolic rules \cite{koza1992genetic,potvin2004generating,brabazon2006evolutionary} to portfolio optimization \cite{chang2000heuristics,brabazon2008introduction} and neuroevolutionary policies \cite{stanley2002evolving,manahov2019high}. While interpretable, these approaches rely on stochastic operators that can yield unstable or brittle strategies in noisy financial environments. Recent co-evolutionary frameworks jointly optimize rules and risk, but remain limited to syntactic search without semantic guidance. AlgoEvolve addresses this gap by replacing random mutation with reasoning-driven LLM transformations, producing logically consistent modifications informed by execution feedback, thus improving stability in non-stationary markets.

\textbf{Meta-Evolution and Automated Search}
Beyond evolving solutions, research explores optimizing the search process itself through self-improving frameworks \cite{real2020automlzero}, meta-learning of optimizers \cite{andrychowicz2016learning}, and co-evolutionary population-based training \cite{jaderberg2017population}. More recently, LLM-driven systems such as PromptBreeder \cite{fernando2023} and OPRO \cite{yang2023opro} demonstrated prompt optimization via self-referential loops in static textual domains. AlgoEvolve extends this paradigm to executable program discovery through a hierarchical architecture in which an outer loop evolves a prompt genome. This enables the autonomous discovery of search heuristics that stabilize exploration and adapt to non-stationary market regimes, mitigating practical failure modes such as strategy degeneration or zero-activity collapse.

\section{Problem Formulation}

We formalize the automated discovery of trading strategies as a \textbf{bi-level, non-stationary program synthesis} problem. The goal is to evolve an agent (strategy) that adapts to shifting data distributions, and simultaneously evolve a search algorithm (prompt) that improves the efficiency of that adaptation.

\subsection{Strategy Representation (The Inner Agent)}
A trading strategy is represented as an executable Python program $f \in \mathcal{F}$, where $\mathcal{F}$ denotes the space of all syntactically valid Python programs expressible under the system’s execution constraints.
Formally, let $\mathbf{x}_t \in \mathbb{R}^d$ be the feature vector at time $t$ (derived from 5-minute OHLCV bars). The strategy computes:
\[
    f(\mathbf{x}_t) \rightarrow \hat{y}_t \in \mathcal{Y}, \quad \mathcal{Y} = \{0,1,2,3,4\} \times \{0,1,2,3,4\}
\]
where $\hat{y}_t$ is a tuple where each component encodes a discrete trading signal over short and long horizons respectively. Unlike parametric models (e.g., Neural Networks) where optimization occurs in weight space $\Theta$, here optimization occurs in the discrete, non-differentiable program space $\mathcal{F}$ \cite{koza1992genetic}.

\subsection{Non-Stationary Objective (Walk-Forward)}
Financial markets are non-stationary, meaning the optimal function $f^\star$ changes over time. We model this using a \textbf{Walk-Forward Validation} protocol \cite{de2018advances}.
Let the lifespan of the system be divided into $K$ temporal epochs. At each epoch $k$, the system has access to a historical window $\mathcal{D}_{train}^{(k)}$. It must produce a strategy $f_k$ to be deployed on the unseen future window $\mathcal{D}_{test}^{(k)}$.

The performance metric $S(f, \mathcal{D})$ is defined as a composite fitness score:
\[
    S(f, \mathcal{D}) = \alpha \cdot \mathcal{R}(f, \mathcal{D}) + (1-\alpha) \cdot \mathcal{C}(f, \mathcal{D})
\]
where $\mathcal{R}$ denotes the \textbf{Total Return} (cumulative Profit and Loss) generated by the strategy, and $\mathcal{C}$ denotes \textbf{Consistency}, a robustness metric measuring the fraction of assets for which the strategy outperforms the median market performance. Here $\alpha$ is a weighting coefficient used to prioritize cross-asset robustness over single-asset outperformance, mitigating the risks of non-stationary distribution shifts. This also serves to avoid the trivial zero-exposure optima observed in preliminary Pareto trials where inactivity satisfies risk constraints perfectly.
The \textbf{Inner Loop} optimization problem at epoch $k$ is to find:
\[
    f_k^\star = \arg\max_{f \in \mathcal{F}} S(f, \mathcal{D}_{train}^{(k)})
\]
The ultimate objective, however, is to maximize the \textit{generalization} to the unseen future: $\sum_{k=1}^K S(f_k^\star, \mathcal{D}_{test}^{(k)})$.

\subsection{Meta-Evolution (The Outer Loop)}
The success of the Inner Loop depends on the search heuristic, governed by the \textbf{Evolver Prompt} $P$. Let $\mathcal{A}(P, \mathcal{D}_{train})$ denote the stochastic optimization procedure (Inner Loop) driven by prompt $P$.
The \textbf{Outer Loop} treats the prompt $P$ as a learnable hyper-parameter. Its goal is to find the prompt $P^\star$ that maximizes the expected Inner Loop performance:
\[
    P^\star = \arg\max_{P \in \mathcal{P}} \mathbb{E}_{\mathcal{D}} \left[ S\left( \mathcal{A}(P, \mathcal{D}_{train}), \mathcal{D}_{test} \right) \right]
\]
This formulation frames the problem as \textbf{Algorithm Discovery}: finding a natural language instruction $P$ that induces an effective search policy in the discrete program space $\mathcal{F}$. This formulation explicitly captures both non-stationary adaptation and self-improving search within a unified optimization framework \cite{hospedales2022metalearning}.

\section{Methodology: The AlgoEvolve Framework}

We propose \textbf{AlgoEvolve} (Figure~\ref{fig:architecture}), a hierarchical evolutionary framework designed to discover trading strategies in non-stationary environments. The system consists of two interacting optimization loops: an \textit{Inner Loop} that evolves executable Python strategies ($f$) and an \textit{Outer Loop} that evolves the natural language instructions ($P$) guiding the search.

\subsection{The Inner Evolutionary Loop}
The Inner Loop functions as a semantic variation-selection engine. Its goal is to approximate the optimal strategy $f^\star$ for a given temporal window. Inner loop execution is controlled by two distinct prompts:
\begin{itemize}
    \item \textbf{The System Prompt (Fixed):} Defines the immutable computational environment, including the dataset structure, allowed Python libraries, and strict I/O constraints (signature enforcement). This prompt is \emph{never} evolved to ensure code validity.
    \item \textbf{The Evolver Prompt (Evolved):} Defines the search heuristic, reasoning style, and creative constraints. This prompt ($P$) is the object of meta-evolution.
\end{itemize}

\subsubsection{Prompt Construction via In-Context Learning}
At generation $t$, the system constructs the input context. To ensure the LLM has access to historical performance data, the system architecture injects the Top-2 Best and Top-2 Worst strategies (i.e., their python code) from the previous generation accompanied by their historical fitness scores into the context window. Contrastive signals from 'Worst' performers enable the LLM to prune ineffective logical branches \cite{romera2023mathematical}, preventing logic collapse during discovery. Crucially, the Evolver Prompt ($P$) determines attention; for example, a prompt might instruct a step-by-step analysis of the divergence between the Best and Worst strategies (strong utilization), or conversely, disregard local optima to propose a radically novel paradigm (zero utilization). This modulates the exploration–exploitation trade-off.

\subsubsection{LLM as a Semantic Mutation Operator}
Unlike random bit-flipping in Genetic Programming, AlgoEvolve uses \textbf{Chain-of-Thought (CoT)} prompting \cite{wei2022chain}. CoT enhances the LLM's capacity for complex symbolic reasoning by enabling it to decompose high-dimensional mutation tasks into intermediate logical steps. The LLM must output a \texttt{<reasoning>} block justifying its proposed changes before outputting the \texttt{<code>} block(s). We make this reasoning-action trajectory a fundamental  requirement rather than an optional heuristic. This ensures mutations are hypothesis-driven (e.g., "The previous strategy over-traded; I will add a volatility filter"). The functional execution of $\text{RunInnerLoop}$ (see Algorithm~\ref{alg:AlgoEvolve}) consists of $N$ generations of semantic evolution. At each iteration $i$, a candidate strategy $f_i$ is synthesized via the mutation operator $\mathcal{M}(f_{best}, \mathcal{L}, P)$. The resulting population is then evaluated on $\mathcal{D}_{train}^{(k)}$ and ranked by the composite fitness $S$, selecting the optimal individual as the parent for the subsequent generation.

\subsubsection{Walk-Forward Evaluation}
Candidate strategies are evaluated using the sliding-window protocol defined in Section 3. The selection pressure is driven by the composite fitness score $S(f)$, balancing Total Return ($\mathcal{R}$) and Consistency ($\mathcal{C}$). Strategies that trigger runtime errors or violate the System Prompt constraints are assigned a fitness of $-\infty$.

\subsection{The Meta-Evolutionary Outer Loop}
While the Inner Loop optimizes the solution, the Outer Loop optimizes the search itself by evolving the Evolver Prompt $P$.

\subsubsection{The Prompt Genome}
We represent the search algorithm as a structured Prompt Genome $G$ consisting of four mutable genes ($\theta$) representing a minimal spanning set for autonomous search \cite{fernando2023}. Each gene $\theta_i$ represents a categorical choice from a curated subspace of natural language instructions. The function $G.\text{build\_prompt}()$ maps these discrete selections into a concatenated, coherent executive directive. This formalization allows the outer loop to perform targeted optimization over specific search behaviors:
\begin{enumerate}
    \item $\theta_{mutation}$: The instruction for modifying code (e.g., "Propose five MINOR variants" vs. "Explore BOLD new paradigms").
    \item $\theta_{focus}$: The creative directive (e.g., "Focus on combining momentum with volatility").
    \item $\theta_{constraints}$: Negative search constraints (e.g., "Do not use look-ahead bias").
    \item $\theta_{reasoning}$: The analytical framework (e.g., "Analyze the learning curve").
\end{enumerate}

\subsubsection{Informed Meta-Mutation}
Standard meta-learning often treats the optimizer as a black box. AlgoEvolve employs \textbf{Informed Meta-Mutation}, where a "Meta-LLM" acts as a research scientist rewriting the genome based on empirical evidence.
After an Inner Loop run, the Meta-LLM receives a \textbf{Performance Report} containing:
\begin{itemize}
    \item \textbf{Learning Curve Trajectory:} To detect stagnation or instability.
    \item \textbf{Failure Rate:} The percentage of candidates that errored (indicating overly restrictive constraints).
    \item \textbf{Champion Anatomy:} The breakdown of Return vs. Consistency.
\end{itemize}
The Meta-LLM is instructed to rewrite exactly one gene to address specific deficiencies identified in the report (e.g., "The search stagnated; rewrite $\theta_{mutation}$ to increase exploration"). By grounding mutations in the Performance Report, this mechanism mitigates meta-learning credit assignment challenges. Rather than performing a stochastic random walk in the prompt space, the Meta-LLM identifies the causal drivers of search failure—such as gradient stagnation or logic collapse—and applies a targeted update to the relevant gene.

\subsubsection{Meta-Crossover}
To combine successful heuristics, we employ uniform crossover in the prompt space. Given two elite genomes, a child genome is created by independently sampling genes from either parent, allowing the system to combine, for example, the \textit{Reasoning Style} of a stable parent with the \textit{Creative Focus} of a high-return parent.

\begin{algorithm}[t]
\caption{AlgoEvolve Meta-Learning Procedure}
\label{alg:AlgoEvolve}
\textbf{Input:} Initial Prompt Population $\mathcal{P}_0$, Meta-Generations $K$
\textbf{Output:} Best Found Prompt Genome $G^\star$
\begin{algorithmic}[1]
    \STATE Initialize population $\mathcal{P} \leftarrow \mathcal{P}_0$
    \FOR{meta-generation $k = 1$ \textbf{to} $K$}
        \STATE $\mathcal{F} \leftarrow \emptyset$, $\mathcal{L} \leftarrow \emptyset$
        \FOR{\textbf{each} Genome $G \in \mathcal{P}$}
            \STATE $P_{text} \leftarrow G.\text{build\_prompt}()$
            \STATE $\text{logs}, \mathcal{H} \leftarrow \text{RunInnerLoop}(P_{text}, \mathcal{D}_{train}^{(k)})$ \COMMENT{$\mathcal{H}$ is the trajectory of candidate programs}
            \STATE $f^\star \leftarrow \text{argmax}_{f \in \mathcal{H}} S(f, \mathcal{D}_{train}^{(k)})$ \COMMENT{Extract strategy maximizing composite fitness $S$}
            \STATE $\mathcal{F}[G] \leftarrow \text{CalculateFitness}(f^\star, \mathcal{D}_{test}^{(k)})$
            \STATE $\mathcal{L}[G] \leftarrow \text{logs}$
        \ENDFOR
        \STATE $\mathcal{P}_{next} \leftarrow \{ \text{argmax}_{G} \mathcal{F}[G] \}$ \COMMENT{Elitism}
        \WHILE{$|\mathcal{P}_{next}| < |\mathcal{P}|$}
            \STATE $p_1, p_2 \leftarrow \text{SelectTop}(\mathcal{P}, \mathcal{F})$
            \STATE $child \leftarrow \text{MetaCrossover}(p_1, p_2)$
            \STATE $\text{report} \leftarrow \text{GenerateReport}(\mathcal{L}[p_1])$
            \STATE $child \leftarrow \text{MetaMutate}(child, \text{report})$
            \STATE $\mathcal{P}_{next}.\text{add}(child)$
        \ENDWHILE
        \STATE $\mathcal{P} \leftarrow \mathcal{P}_{next}$
    \ENDFOR
    \STATE \textbf{return} $\text{argmax}_{G} \mathcal{F}[G]$
\end{algorithmic}
\end{algorithm}

\section{Experiments}

\subsection{Experimental Setup}

\subsubsection{Market Environment and Feature Representation}
We evaluate AlgoEvolve on the NUMIN \footnote{\url{https://pypi.org/project/numin/}} platform, a publicly available \textit{intraday} paper trading environment with open programmatic access, providing anonymized multi-asset equity market data and programmatic strategy evaluation via a Python SDK. The dataset consists of over 200 trading days of discretized 5-minute data that includes OHLCV candles, standard technical indicators, as well as 5 and 10 candle returns, normalized by the first close price of each day. Each trading day is represented by 150 time steps, where rows 0–74 correspond to historical prior-day context and rows 75–149 represent the active trading session. Each day's data is for a different set of 5 (obfuscated) stock symbols. Normalized and obfuscated market data ensures numerical stability and prevents leakage. The back-testing SDK takes as input a strategy function and computes its returns for a given set of days as follows:  The strategy function is provided data prior to a given 5-minute tick (omitting returns that are unavailable at that time), and decides to buy, sell or hold a stock. Each position exits 10 candles after entry, and a single open position per stock symbol is enforced. A fixed $0.1\%$ (10bps) transaction plus slippage cost is enforced to prioritize high-conviction signals over noise.

Strategy agents are required to output predictions for the dual multi-horizon targets defined in Section 3.1. We maintain the same five-class discrete scale $\mathcal{Y}$ for all discovered logic, ensuring that the semantic mutation process remains grounded in the established numerical stability constraints.

\subsubsection{Model Configuration and Hyperparameters}
We employ a heterogeneous LLM architecture to decouple the high-latency reasoning required for prompt optimization (Gemini Pro) from the high-throughput generation required for strategy discovery (Gemini Flash). Critically, since the Gemini Flash 'generator' remains constant across all ablation stages, the performance gains observed in Stage 4 are isolated to the quality of the evolved instructions rather than a shift in model-tier capacity. The Outer Loop utilizes the Pro family for its superior reasoning capabilities and long-context window, while the Inner Loop leverages the Flash family for rapid iterative refinement. We set the consistency weighting coefficient empirically as $\alpha = 0.7$, as defined in Section 3.2.

\subsection{Evaluation Metrics and Protocols}
To rigorously assess \textbf{AlgoEvolve}, we utilize a multi-dimensional evaluation framework that accounts for absolute profitability, risk exposure, and structural stability across non-stationary regimes.

\subsubsection{Performance Metrics}
We evaluate strategies using standard financial performance measures that jointly assess return potential and downside risk: 
\textbf{Average Daily Profit and Loss (Avg. PnL)}, the arithmetic mean of daily portfolio returns; 
\textbf{Sharpe Ratio}, measuring the consistency of alpha generation via risk-adjusted returns; 
\textbf{Return Volatility}, representing the variability of portfolio returns; and 
\textbf{Maximum Drawdown (MDD)}, quantifying the worst-case cumulative capital loss from a local peak to a subsequent trough. 
All Sharpe ratios are reported on a daily basis unless otherwise stated; annualized values use $S_{\text{ann}} = S_{\text{daily}} \times \sqrt{252}$ for industry-standard comparison.

\subsubsection{Temporal Evaluation Protocols}
We assess the framework across two paradigms: \textbf{Fixed (Disjoint) Window}, where the evaluation window $W$ shifts by its total length (e.g., 13 days) to establish long-term logic stability; and \textbf{Sliding Window}, where the window shifts by a single day. The latter provides high-frequency feedback, enabling the mutation operator to adapt to immediate market regime shifts. Together, these protocols provide a view of both economic value and robustness.

Unlike static ML benchmarks, intraday markets exhibit frequent regime shifts. We prioritize rapid adaptation under continual distribution shift over asymptotic convergence, utilizing high-frequency rolling windows to assess semantic mutation.

\subsection{Results and Ablation Study}
In this section, we analyze the performance of the \textbf{AlgoEvolve} framework through an incremental ablation study. By isolating the impact of temporal windowing, feedback granularity, and meta-evolutionary optimization, we demonstrate how the system systematically overcomes market non-stationarity.

Prior to the longitudinal study, preliminary calibration experiments were conducted to evaluate the impact of temporal windowing. Comparing 13-day disjoint and 5-day sliding windows, we established that a high-frequency, 1-day feedback protocol optimizes for recency bias and minimizes drawdown. This configuration serves as our 'Standard Evol' baseline (Mean: $0.104\%$, Daily Sharpe: $0.36$, MDD: $0.42\%$), representing the limits of inner-loop semantic evolution without the adaptive guidance of the meta-evolutionary outer loop.

\subsubsection{Meta-Evolutionary Optimization}

\begin{figure}[t]
    \centering
    \includegraphics[width=\columnwidth]{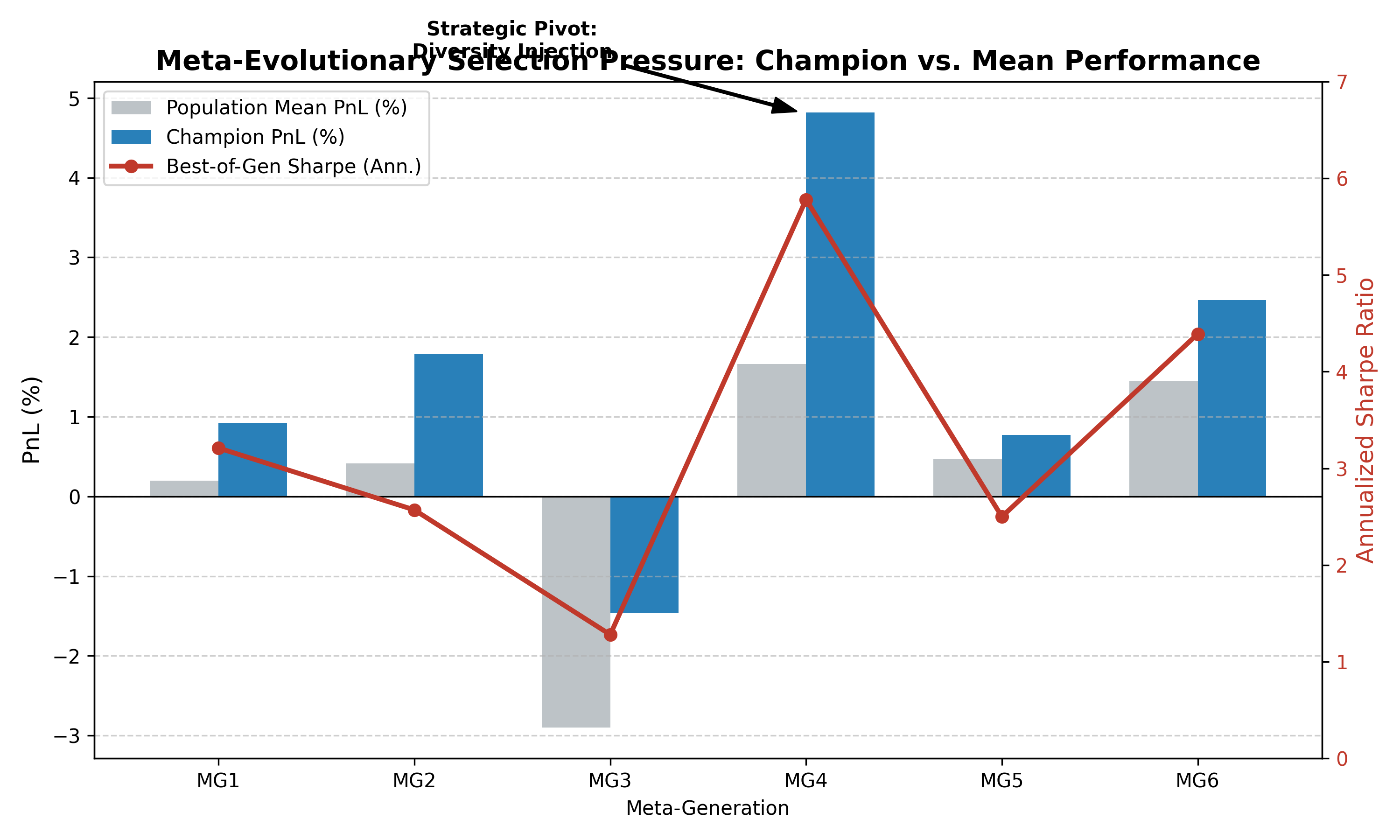}
    \caption{Evolution of the Prompt Genome. Meta-evolutionary selection identifies elite heuristics achieving an \textbf{annualized Sharpe of 5.60} against a population mean of 1.21, representing a \textbf{363\% improvement} in risk-adjusted search efficiency over six meta-generations.}
    \label{fig:prompt_evolution}
\end{figure}

AlgoEvolve successfully navigates non-stationary search spaces through autonomous self-correction, exhibiting a significant \textbf{selection advantage} (Fig.~\ref{fig:prompt_evolution}). If the synthesis was unreliable, resulting in incorrect code being generated, our system treated the prompt genome as having $-\infty$ fitness.

\begin{figure}[t]
    \centering
    \includegraphics[width=\columnwidth]{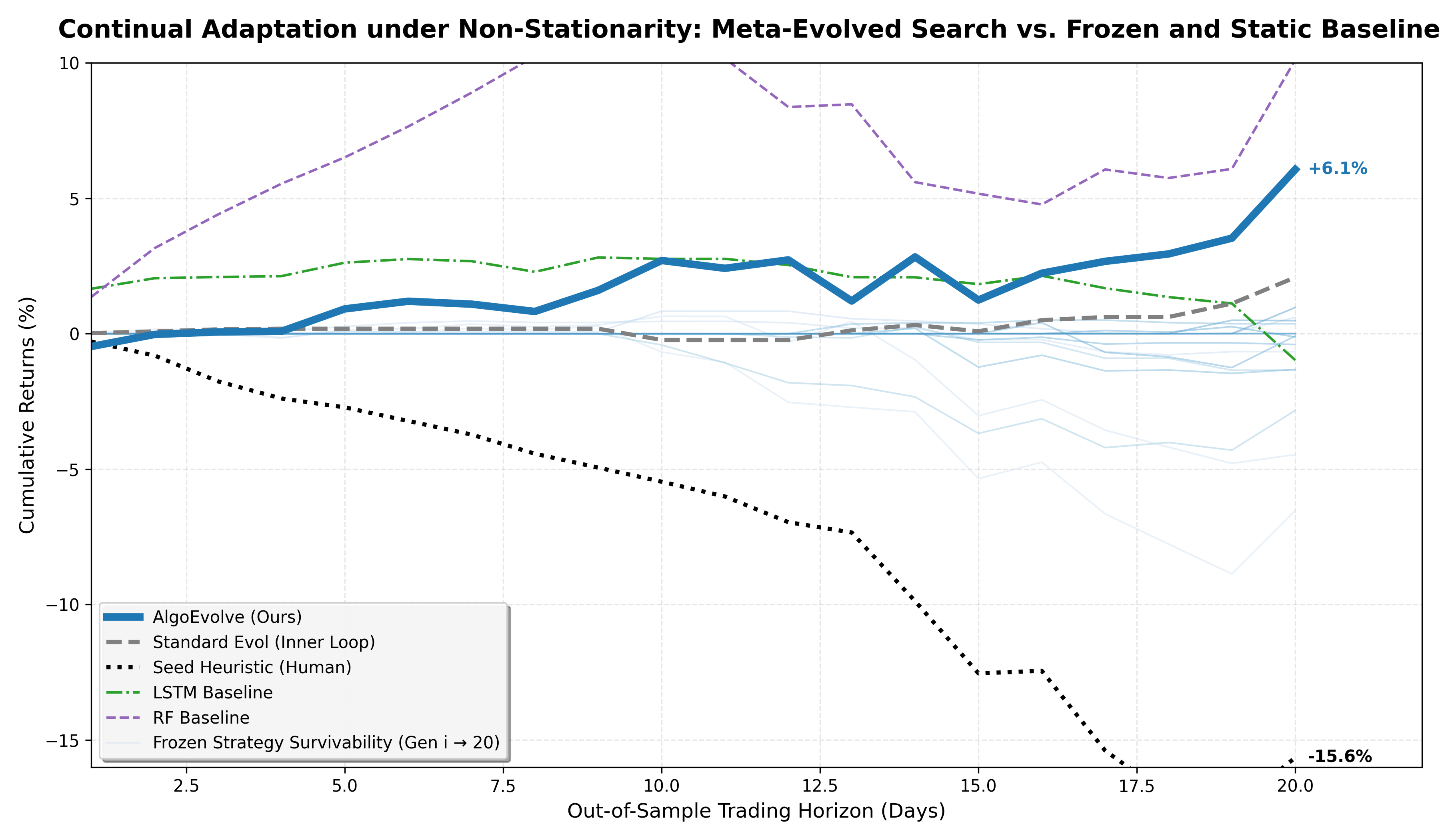}
    \caption{Equity curves under a rolling walk-forward evaluation. AlgoEvolve (Outer Loop) maintains a stable upward trajectory, while static and parametric baselines exhibit high variance and suffer drawdowns following the regime shift after Day 10.}
    \label{fig:equity_curves}
\end{figure}

\subsection{Comparative Performance Analysis}
To assess robustness under market non-stationarity, we evaluate \textbf{AlgoEvolve} against four standardized baselines: a static \textit{Seed} heuristic, a single-level \textit{Standard Evol} ablation, and two tuned parametric models (\textit{RF}, \textit{LSTM}). The parametric baselines were trained on a 95-day dataset (June--Oct 2024) utilizing 59 market features: the \textbf{Random Forest (RF)} employs 250 estimators with a depth of 8 over a 30-candle window, while the \textbf{LSTM} utilizes a 2-layer architecture (128 hidden units) with a 20-step temporal lookback. Additionally, we include ``Frozen'' inner-loop champions ($Gen_{i} \rightarrow 20$) to measure the rate of strategy decay in the absence of meta-adaptation (Table~\ref{tab:performance}, Fig.~\ref{fig:equity_curves}).

\begin{table}[t]
\centering
\small
\setlength{\tabcolsep}{3pt}
\begin{tabular}{lrrrr}
\toprule
Method & Mean (\%) & Volatility (\%) & Ann. Sharpe & Max DD (\%)\\
\midrule
Seed Heuristic         & -0.78 & 1.06 & -11.75 & 17.56 \\
Standard Evol          &  0.10 & 0.29 &  5.71 &  0.42 \\
LSTM                   & -0.05 & 0.68 & -1.11 &  3.79 \\
RF                     &  0.51 & 1.52 &  5.24 &  7.27 \\
\textbf{AlgoEvolve} 
                       & \textbf{0.30} & \textbf{0.94} & \textbf{5.08} & \textbf{1.59} \\
\textbf{AlgoEvolve (6 MG)} & \textbf{0.31} & \textbf{0.88} & \textbf{5.60} & \textbf{1.59} \\
\bottomrule
\end{tabular}
\caption{Comparative performance of strategy discovery methods across a 20-day benchmark and a 30-day longitudinal (6 MG) evaluation. AlgoEvolve demonstrates superior risk-adjusted alpha (Sharpe: 5.60) and autonomous resilience to non-stationarity. While parametric baselines (RF, LSTM) exhibit high-variance decay and severe drawdowns (7.27\%), our meta-evolutionary framework identifies elite heuristics that maintain capital preservation (Max DD: 1.59\%) while consistently capturing positive mean returns. AlgoEvolve's Sharpe here corresponds to the elite Prompt Genome selected under evolutionary pressure, while the population-level mean Sharpe remains $\approx 1.21$; our claims are therefore comparative and selection-based rather than reflective of average deployable performance.}
\label{tab:performance}
\end{table}

\textbf{Performance Benchmarking.} Table \ref{tab:performance} shows \textbf{AlgoEvolve (6 MG)} achieves the optimal risk-return profile. The \emph{Seed Heuristic} fails catastrophically ($-15.65\%$ loss), while the \emph{Standard Evol} baseline reveals single-loop limits: high stability (MDD $0.42\%$) but muted alpha ($0.10\%$ mean); it lacks the structural verticality required to capitalize on high-alpha events. Conversely, our bi-level architecture unlocks significant ``verticality,'' increasing mean daily returns to $0.31\%$---a \textbf{3$\times$ improvement}---with an institutional-grade \textbf{5.60 Sharpe}. 

\textbf{Baseline Fragility.} Parametric models exhibit regime-shift decay: the \emph{RF Baseline} suffers high volatility and severe drawdown ($7.27\%$), while the \emph{LSTM} fails to extract signal from intraday noise. AlgoEvolve’s symbolic, rule-based heuristics act as a natural regularizer, mitigating the high-frequency overfitting inherent in black-box models.

\textbf{Adaptation Dynamics.} Superior performance is driven by the outer loop's strategic pivots. As seen in Fig.~\ref{fig:equity_curves}, the \emph{Standard Evol} curve plateaus, a trend mirrored by \textbf{Frozen Strategy fossils} ($\mathrm{Gen}_{i} \rightarrow 20$) which decay as distributions shift. AlgoEvolve overcomes this ``alpha silence'' by identifying the \textbf{MG 3 drawdown} ($-1.59\%$) and autonomously injecting search diversity. This triggered a \textbf{V-shaped recovery} in MG 4, capturing a $2.53\%$ single-day alpha event, demonstrating how design-time reasoning can enable structural adaptation beyond runtime parametric inference.

\section{Qualitative Analysis}

\subsection{Inner-Loop Baseline Analysis}

Preliminary calibration (13-day window) established the framework’s ability to autonomously pivot from failed trend-following to mean-reversion logic, validating the high-frequency protocol. Subsequent ablative analysis of the ``Standard Evol'' (Inner-Loop only) trajectory reveals a distinct three-phase decay pattern.

\paragraph{Phase 1: Rapid Heuristic Convergence (Gens 1--4).}
In the initial generations, the LLM successfully synthesized a foundational suite of technical indicators. By Generation 1, the system discovered the utility of \textbf{RSI/CMO momentum filters} and \textbf{SMA trend confirmation}. Reasoning logs show the model attempting to ``create high-granularity signals'' by aggregating multiple binary conditions. However, the logic remained shallow, relying on fixed thresholds (e.g., \texttt{buy\_threshold = 4.0}) that lacked sensitivity to intraday volatility regimes.

\paragraph{Phase 2: The ``Zero-Trade'' Stagnation Trap (Gens 5--12).}
The most significant finding in the baseline study was a prolonged period of \textbf{semantic stagnation}. During these eight generations, the system produced a terminal return of $0.0\%$. 
\begin{itemize}
    \item \textbf{Logic Collapse:} Reasoning traces indicate the LLM attempted to ``enhance robustness'' by adding increasingly complex conditional layers. This resulted in unintentional \textbf{over-regularization} of the trading logic.
    \item \textbf{Syntactic Over-Smoothing:} Because the Evolver Prompt remained static, the LLM became trapped in a loop of minor threshold refinements. The resulting programs were so conservative that they failed to trigger a single trade across validation assets, identifying a state of \textbf{``Alpha Silence''}: when executable logic could not be generated by the provided instructions.
\end{itemize}

\paragraph{Analysis: Threshold Fragility and Bi-Level Necessity.}
As illustrated in Table \ref{tab:logic_evolution}, the baseline search eventually attempts to recover from stagnation through \textbf{Threshold Fragility} (Gen 20)—performing stochastic parameter tuning on aggressive 'overextension' rules rather than structural innovation. This progression confirms that under static prompts, LLMs as mutation operators are prone to local optima convergence. A meta-evolutionary layer is therefore required to mutate the search instructions themselves, forcing the synthesizer to abandon redundant filters and discover the regime-adaptive architectures seen in the full AlgoEvolve framework.

\begin{table}
\centering
\small
\begin{tabularx}{\columnwidth}{lX}
\toprule
\textbf{Epoch} & \textbf{Discovered Heuristic Logic} \\
\midrule
Gen 1 & \texttt{if RSI\_14 > 70: return SELL} (Naive Threshold) \\
Gen 10 & \texttt{if (RSI\_14 > 75 \&\& Close\_n > SMA\_20): return 2} (Over-filtered/Zero-trade) \\
Gen 20 & \texttt{score = $\sum (w_i \cdot \phi_i)$; if score > $\tau$: return 4} (Multi-factor Confluence) \\
\bottomrule
\end{tabularx}
\caption{Evolution of symbolic logic. The system autonomously transitioned from naive thresholds to weighted multi-factor scoring architectures.}
\label{tab:logic_evolution}
\end{table}

% \paragraph{Phase 3: Erratic Re-acceleration and Threshold Fragility (Gens 13--20).}
% A late-stage attempt at recovery occurred in Generation 13 ($+0.35\%$ return), where the model shifted focus to ``Dynamic Slope-Momentum.'' However, without the Outer Loop to recalibrate the search philosophy, the inner loop struggled with \textbf{Threshold Fragility}. The performance spike in Generation 20 ($+0.96\%$) was driven by aggressive ``Extreme Overextension'' logic. This logic was structurally similar to the failed logic of Generation 10 ($-0.41\%$), suggesting the inner loop was performing stochastic threshold optimization rather than evolving a fundamentally superior paradigm.

% \paragraph{Synthesis: The Case for Bi-Level Evolution.}
% The baseline demonstrates that \textbf{LLMs as mutation operators are prone to local optima convergence}. When a static prompt instructs a model to ``improve the strategy,'' the model tends to add redundant filters until the strategy stops trading entirely (the \textbf{Zero-Trade Trap}). This confirms that a second level of evolution---one that can mutate the search instructions themselves---is necessary to break stagnation and force the discovery of truly regime-adaptive programs.

\subsection{Meta-Evolutionary Dynamics}

In contrast, as illustrated in Fig.~\ref{fig:prompt_evolution}, the \textbf{AlgoEvolve} meta-trajectory followed \textbf{three distinct evolutionary phases} characterized by initial optimization, a regime crisis, and a strategic recovery breakthrough. This corresponds to three distinct Prompt Genome Eras:

\paragraph{Era 1: The Refinement Paradigm (MG1--MG2).}
In the initial phase, the meta-population was dominated by ``Convergent DNA'' (e.g., \texttt{prompt\_12e806c5}), characterized by instructions to \textit{``Propose MINOR variants... small, logical refinements.''} This search style was highly effective in stable regimes, identifying a ``Confluence Filter'' that successfully combined RSI and Bollinger Bands. In MG2, the system autonomously discovered the \textbf{Hierarchical VWAP Filter} (\texttt{prompt\_b1c7b1ac}). By mandating that all entry logic be gated by VWAP, the Meta-LLM synthesized a regime-aware search instruction. This prevented inner-loop models from trading against the dominant trend, leading to a cumulative PnL of $+1.79\%$.

\paragraph{Era 2: The Resilience Pivot (MG3--MG4).}
The system encountered a performance plateau due to a market regime shift in MG3, where previously optimized logic failed (Champion loss: $-1.59\%$, Population mean: $-2.90\%$). Detailed logs indicate the inner-loop models were over-fitting to prior volatility. In response, the Meta-LLM executed a \textbf{Strategic Diversity Injection} in MG4 (\texttt{prompt\_94876739}). As shown below, the evolved $\theta_{focus}$ gene shifted the search paradigm to overcome the alpha plateau: e.g., \textit{``Propose structurally diverse archetypes; explore non-linear indicator combinations and inverse volatility signals to bypass the current zero-trade stagnation.''} This transition provides empirical evidence of the system's capacity to autonomously detect heuristic failure and execute a structural pivot, resulting in a recovery that captured a 2.53\% alpha event in MG4.

\paragraph{Era 3: The Complexity Ceiling (MG5--MG6).}
The final phase of the longitudinal study identifies a complexity saturation point at the frontier of autonomous synthesis. As meta-instructions evolved toward high-density logic, the framework encountered its first syntactic boundary in the final generation. Crucilly, this served as a validation of the system's robustness-aware selection: while a subset of the population errored(no valid asset results), the robustness filters successfully quarantined these candidates, assigning them $-\infty$ fitness. Despite this tail-end complexity, the search remained productive, with the champion lineage remaining free of invalid code and capturing a significant $+2.50\%$ alpha event. This minor blip at the experimental frontier demonstrates that the framework is capable of safe continual discovery, prioritizing execution integrity over speculative complexity.

\begin{table}[t]
\centering
\small % Sets font to 9pt (IJCAI Section 3 compliant)
\setlength{\tabcolsep}{6pt} % You can increase padding now that there's room
\begin{tabular}{llc}
\toprule
Meta-Gen & Evolutionary Breakthrough & Impact (PnL) \\
\midrule
MG1 & \textbf{Logical Confluence}: Combined momentum/trend. & $+0.91\%$ \\
MG2 & \textbf{Hierarchical Gating}: VWAP mandatory filter. & $+1.79\%$ \\
MG3 & \textbf{Regime Shift}: Static logic fails in new regime. & $-1.59\%$ \\
MG4 & \textbf{Diversity Portfolio}: Non-linear/inverted logic. & $+4.82\%$ \\
MG6 & \textbf{Complexity Ceiling}: Synthesizer limits reached. & $+2.46\%$ \\
\bottomrule
\end{tabular}
\caption{Qualitative Heuristic Evolution. The table maps evolved search instructions to their quantitative impact, highlighting the autonomous pivot in MG4.}
\label{tab:meta_evol_path}
\end{table}

\section{Conclusion and Outlook}
AlgoEvolve provides a bi-level evolutionary framework that reframes LLMs as symbolic architects capable of autonomous strategy discovery. We demonstrated the system's ability to navigate non-stationary market regimes, achieving a peak annualized Sharpe ratio of 5.60. To address the constraints identified during our study, future work can explore \textbf{Modular Program Synthesis} to overcome the observed complexity ceiling by evolving discrete, reusable \textit{sub-routines}.

\bibliographystyle{named}
\bibliography{ijcai26}

\end{document}